\documentclass[letterpaper, 10 pt, conference]{ieeeconf}  

\IEEEoverridecommandlockouts                              
\usepackage{subcaption}
\usepackage{mwe}
\usepackage{graphics} 
\usepackage{bm}
\usepackage{siunitx}
\usepackage{relsize}
\usepackage{amsmath} 
\usepackage{amssymb}  
\usepackage[ruled,vlined]{algorithm2e} 
\usepackage{booktabs}
\usepackage{siunitx}
\usepackage{multirow}
\usepackage{makecell}
\usepackage{subcaption} 
\usepackage{hyperref}
\usepackage[usenames, dvipsnames,table,xcdraw]{xcolor}
\usepackage{pifont}

\usepackage{tabularx}
\UseRawInputEncoding

\usepackage{amssymb}
\usepackage{array} 
\usepackage{arydshln}
\usepackage[table]{xcolor}
\usepackage[dvipsnames]{xcolor}
\usepackage{etoolbox}
\usepackage{todonotes}
\newcolumntype{s}{>{\color{blue}}c}

\usepackage[capitalise]{cleveref}
\usepackage{bm}
\usepackage{mathtools}

\title{\LARGE{\bf{WOLF-VLA: Whole-Body Humanoid Optimal Locomotion Framework for Vision-Language-Action Learning}}}
\author{Melya Boukheddimi$^{1*}$,Omar Adjali$^{2*}$, Daniel Sontag$^{2,3}$, and Frank Kirchner$^{1,4}$
\thanks{This work was supported by the German Federal Ministry of Research, Technology and Space (BMFTR) through the ActGPT project (Grant No. 01IW25002), the RV-NI-2024–2029-K-IML project (Grant No. 25361), the Amenable project (Grant No. 16IW26002), and the hessian.AI Service Center (Grant No. 16IS22091). Additional support was provided by the hessian.AI Innovation Lab, funded by the Hessian Ministry for Digital Strategy and Innovation (Grant No. S-DIW04/0013/003), the Robotics Institute Germany (RIG), the Lower Saxony Ministry of Science and Culture (MWK) through the zukunft.niedersachsen program, and the Endowed Chair of Artificial Intelligence (AI) at the University of Oldenburg.}
\thanks{{$^{*}$Both authors contributed equally to the paper}}
\thanks{$^{1}$Robotics Innovation Center, DFKI, 28359 Bremen, Germany.}%
\thanks{$^{2}$Interactive Machine Learning, DFKI, 26129 Oldenburg, Germany}%
\thanks{$^{3}$University of Oldenburg, 26129 Oldenburg, Germany.}%
\thanks{$^{4}$AG Robotik University of Bremen, 28359 Bremen, Germany.}%
\thanks{Corresponding authors:
\href{mailto:melya.boukheddimi@dfki.de}{melya.boukheddimi},
\href{mailto:omar.adjali@dfki.de}{omar.adjali}\texttt{@dfki.de}.}
}

\begin{document}
\newcommand{\robotName}{RH5V2}

\newcommand{\mvec}[1]{\bm{#1}}
\newcommand{\vc}[1]{\mathbf{\mathbf{#1}}}

\newcommand{\q}{\textbf{q}}
\newcommand{\dq}{\dot{\q}}
\newcommand{\ddq}{\ddot{\q}}


\newcommand{\Mass}{\mathbf{M}}
\newcommand{\Bias}{\mathbf{b}}
\newcommand{\Gravity}{\mathbf{g}}
\newcommand{\Force}{\mathbf{\lambda}}
\newcommand{\Torque}{\mathbf{\tau}}
\newcommand{\Jac}{\mathbf{J}}

\newcommand{\BIN}{\begin{bmatrix}}
\newcommand{\BOUT}{\end{bmatrix}}

\newcommand{\sref}[1]{Sec~\ref{#1}}
\newcommand{\eref}[1]{(\ref{#1})}
\newcommand{\fref}[1]{Fig.~\ref{#1}}
\newcommand{\tref}[1]{Table~\ref{#1}}
\newcommand{\equationref}[1]{Equ.~\ref{#1}}
\newcommand{\state}{\mathbf{x}}
\newcommand{\ctrl}{\mathbf{u}}
\newcommand{\dynsys}{\mathbf{f}}

\newcommand{\qTr}{\underline{\q}}
\newcommand{\dqTr}{\underline{\dq}}
\newcommand{\ddqTr}{\underline{\ddq}}
\newcommand{\TorqueTr}{\underline{\Torque}}

\newcommand{\costl}{l}
\newcommand{\dts}{\Delta t_s}
\newcommand{\st}{\text{subject to}}

\maketitle
\thispagestyle{empty}
\pagestyle{empty}
\begin{abstract}
Vision-Language-Action (VLA) models have recently demonstrated strong generalization in robotic manipulation, yet their applicability to whole-body, contact-rich humanoid locomotion remains severely underexplored due to data scarcity, the absence of dynamically consistent demonstrations, and the difficulty of encoding optimality and safety in learning-based pipelines. This work introduces a unified framework WOLF-VLA that integrates whole-body optimal-control motion synthesis with large-scale multi-modal dataset to train VLAs capable of generating humanoid locomotion policies directly from natural-language instructions. We construct a comprehensive dataset of dynamically feasible humanoid trajectories across six locomotion-related task families, each parameterized by environmental variations, object colors, placements, and visual distractors. 
We train a VLA model using the collected joint trajectories, ego-centric visual observations and natural language instruction, yielding a policy that exhibits strong reasoning and robustness to initial-condition variability, and competitive performance across several  tasks and environment settings. A systematic ablation study demonstrates the impact of each modality on the model performance. 
The full dataset, model checkpoints, and benchmarking simulation suite will be openly released, establishing a reproducible dynamically consistent benchmark for whole-body humanoid locomotion rich VLA control and enabling future research in scalable transfer of instruction-driven locomotion policies. 
\end{abstract}
\section{Introduction}
\label{sec:intro}
Over the past decades, motion generation and control in robotics, particularly for complex legged robots such as humanoids and quadrupeds, has relied heavily on Optimal Control (OC) methods \cite{MelyaLift} 
\cite{mastalli2020direct}. The dominant research trend combined OC techniques with whole-body control for stabilization and Model Predictive Control (MPC) for motion adaptation and robustness \cite{anymalmpc}. A large portion of the community focused on refining these tools in order to achieve stable and reliable real-world deployment.
In parallel, a different line of research emerged and began producing strong results on physical systems. Leveraging learning-based 
\begin{figure}[h]
	\centering
	\includegraphics[width =0.5\textwidth]{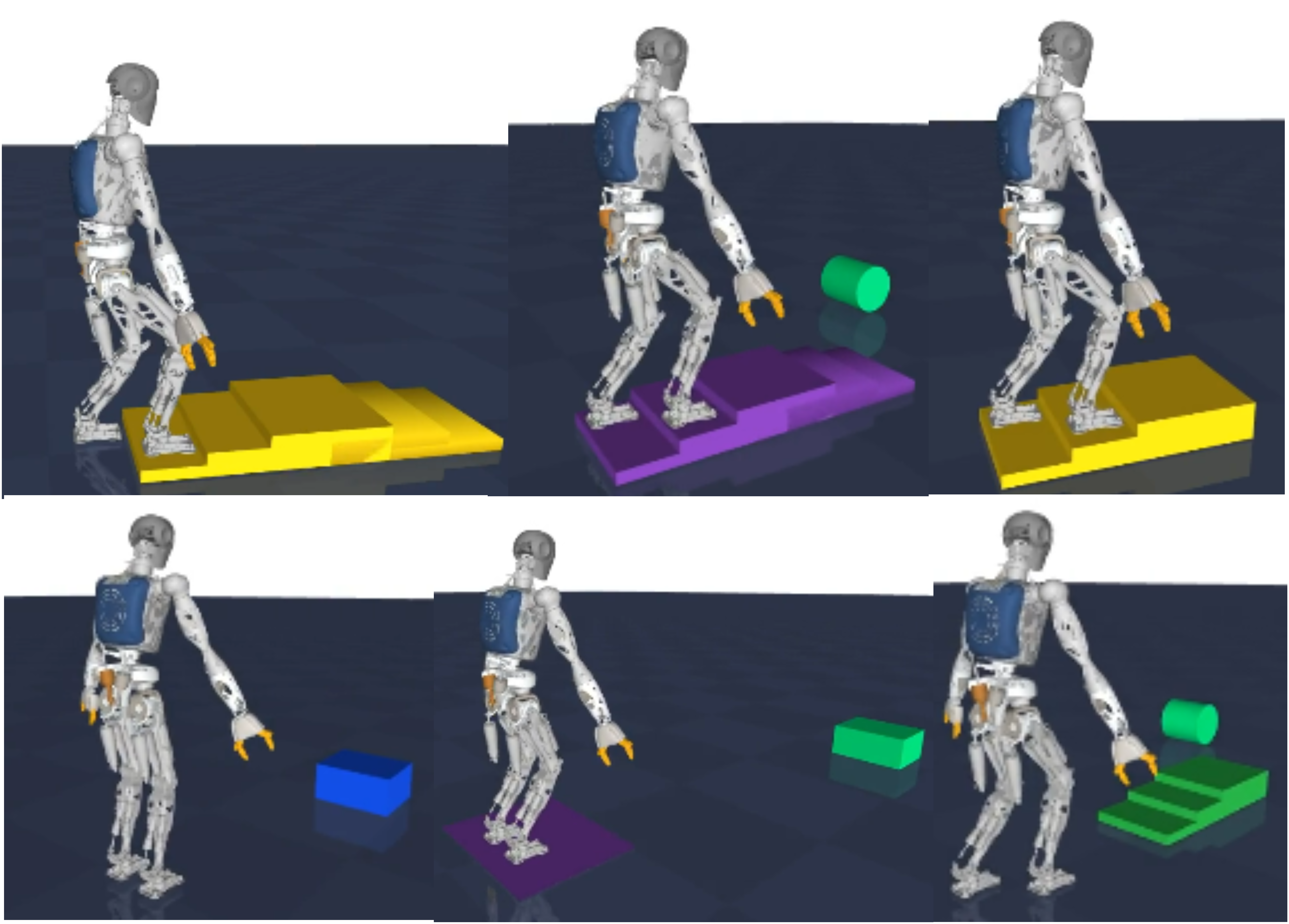}
\caption{Illustrative snapshots of evaluation cases highlighting the breadth of task and environmental variability.}
	\label{fig:open}
     \vspace{-0.8cm}
\end{figure} 
 paradigms \cite{tong2024advancements}, long recognized in theory for their expressive power,  this community demonstrated highly adaptive and versatile behaviors enabled by algorithms such as Reinforcement Learning (RL) \cite{smith2022walkparklearningwalk, radosavovic2024real}, Imitation Learning (IL) \cite{peng2018sfv, peng2021amp}, and deep learning  \cite{muzio2022deep}.
At the same time,  breakthroughs in large-scale ML, originally driven by text and image domains, introduced multi-modal ML models to robotics. Vision-Language-Action (VLA) models, supported by modern GPU computation and large-scale training strategies, have recently achieved real-time deployment on robotic hardware with promising levels of generalization and adaptability.
One of the first major advances in this area came from the Physical Intelligence group \cite{black2024} \cite{intelligence2025pi05visionlanguageactionmodelopenworld}. Their contributions sparked a wave of follow-up work, opening a completely new avenue for robot motion generation and control. This line of research is paving the way for robots that are more interactive and adaptive than ever, integrating multi-modal perception with versatile control policies. These methods endow robots with heightened reactivity, adaptability, and the capacity for rich interaction in dynamic environments, and because they are model-free, they can be seamlessly transferred to a wide range of robotic platforms. This combination of capabilities fills several long-standing missing pieces in the field of robotics, that could enable the next stage of deployment not only in industrial settings but, for the first time, potentially in everyday life.
Despite these advances, several limitations remain. Most existing VLAs are demonstrated on relatively simple robotic platforms, typically fixed-base manipulators, systems with low kinematic complexity, or tasks that do not require highly dynamic or high-inertia motion. Their applicability to whole-body, highly dynamic, multi-contact systems such as humanoid locomotion has only very recently begun to be explored, and current contributions remain very limited.
Another major limitation concerns the datasets required for training. Many approaches rely on teleoperation to control robotic arms and record the resulting motions and images, which are then labeled. Others collect data using motion-capture systems while operating the robot for hours. These procedures are costly, time-consuming, and restricted to teams with access to the necessary hardware.
Additionally, the collected motions typically lack the notion of optimality. The robot may walk or reach a target, but its behavior does not encode energy efficiency, minimal joint motion, or other optimality criteria that classical model-based OC methods naturally enforce.
More importantly, these data-driven methods do not inherently guarantee safety with respect to joint position, velocity, or torque limits. While teleoperated kinematic robotic arms face relatively low risk, the situation is different for highly dynamic legged robots such as quadrupeds or humanoids. In these cases, safety must be ensured through additional control modules, usually inserted at the sim-to-real transfer, since model-free policies do not provide the safety and optimality guarantees traditionally offered by OC-based approaches.
\paragraph*{Contributions}
In order to address the limitations of existing humanoid locomotion resources most notably the lack of large-scale whole-body humanoid datasets and the absence of motion data that are physically optimal, dynamically consistent, and contact-coherent, we propose a framework that generates whole-body humanoid locomotion exclusively through optimal-control problem (OCP) resolution. This framework produces inherently optimal trajectories that improve the quality of behaviors learned from them. Using the dataset produced by this framework, we train and evaluate a VLA model that achieves improved performance and robustness across diverse locomotion tasks.
The primary contributions of this work are summarized as follows:
\begin{itemize}
    \item WOLF-VLA-dataset a large-scale dataset of whole-body humanoid locomotion covering forward walking, stair climbing, side gait, squatting, and robot rotation.
    
    \item OC-based generated trajectories that are inherently optimal, smooth, feasible, and physically consistent motions suited for high-quality policy learning.
    
    \item A new benchmark for whole-body humanoid locomotion within VLA, enabling evaluation of challenging tasks.
    
    \item WOLF-VLA-model a strong VLA model baseline fine-tuned and evaluated on the proposed benchmark.
    
    
    \item Systematic ablation study assessing: the influence of different control modalities on task performance and the robustness under injected perturbations to quantify the stability and reliability of the model.

\end{itemize}
\paragraph*{Organization}
The paper is organized as follows. \sref{sec:related_work} reviews the most relevant state-of-the-art
work on VLA models and datasets, discussing
their respective strengths and limitations. \sref{sec:methodology} introduces the pipeline methodology followed in this work, while \sref{sec:implementation} describes the implementation details and the metrics used. \sref{sec:results} reports and discusses the obtained results. Finally, \sref{sec:conclusion} highlights the limitations and outlooks of our study and concludes the paper.
\section{Related Work}
\label{sec:related_work}
Recent advances in robot control have been strongly influenced by the emergence of multi-modal policy frameworks driven by large generative models. These frameworks typically integrate one or more high-level modalities language (L), vision (V), or a combination of both (VL) together with an action model that grounds these modalities in the robot's embodiment. While large language models (LLMs) and vision-language models (VLMs) have already matured and provide powerful policy priors, the integration of an explicit action model remains a key innovation for robotics. 
The action model encodes the robot’s physical embodiment, including its control spaces, and state transitions. By coupling this embodiment-aware component with multi-modal policies, robots gain the capacity to interpret natural language (NL) instructions, perceive complex environments, and generate contextually grounded actions. This integration significantly broadens the design space of robot capabilities: NL interfaces improve human-robot interaction, visual grounding allows adaptation to unstructured environments, and action grounding ensures robot reactivity.
\subsubsection*{Multi-Modal Robot Control}
Early multi-modal robot control was driven by RT-2 ~\cite{brohan2023rt2visionlanguageactionmodelstransfer}, a 55B-parameter VLA model trained on Internet-scale multi-modal data and robot demonstrations. RT-2 introduced semantic reasoning and achieved strong generalization, but remained limited to fixed-base manipulation. OpenVLA~\cite{kim2024openvlaopensourcevisionlanguageactionmodel}, a 7B-parameter model trained on 970k demonstrations from the Open X-Embodiment dataset~\cite{o2024open}, improved RT-2 by 16\% on standard manipulation benchmarks while providing open-sourced checkpoint. However, similar VLA approaches~\cite{etukuru2024robotutilitymodelsgeneral,driess2023palmeembodiedmultimodallanguage} 
remain constrained to manipulation setups and to their reliance on discrete next action prediction.
Recent models target lighter architectures and improved embodiment generalization. Pi\textsuperscript{0} \cite{black2024} introduced a 3B-parameter VLA with action-flow modeling for broader transfer across embodiments, while FLOWER~\cite{reuss2025flowerdemocratizinggeneralistrobot} achieved competitive performance across 190 tasks and 10 benchmarks using a compact 1B-parameter architecture thanks to its intermediate modality fusion and Global-AdaLN conditioning. GROOT-N1~\cite{nvidia2025gr00tn1openfoundation} employs a 2.2B-parameter diffusion transformer for action modeling, enabling single-arm and bi-manual humanoid manipulation from mixed real, simulated, and human-video datasets. 
Similarly, $\pi_{0.5}$\cite{intelligence2025pi05visionlanguageactionmodelopenworld} extends $\pi_{0}$ through hybrid multi-modal co-training to further enhance embodiment adaptability. Several lightweight controllers have also emerged. ACT-1 \cite{zhao2023learningfinegrainedbimanualmanipulation} is an 
∼80M-parameter vision-action model with strong data efficiency, while SmolVLA \cite{shukor2025smolvla} uses multi-camera observations and NL instructions within only 450M-parameter policy-learning framework.
Beyond manipulation, recent efforts address whole-body VLA control. Radosavovic et al.~\cite{radosavovic2024humanoid} formulate humanoid locomotion as next-token prediction using MoCap and video supervision.
LeVERB~\cite{xue2025leverb} learns a latent space from synthetic demonstrations and executes it via a whole-body RL controller, while WholeBodyVLA~\cite{jiang2025wholebodyvla} uses action-free egocentric videos for latent-action pretraining prior to RL. SENTINEL~\cite{wang2025sentinel} directly predicts action chunks from language and proprioception using large-scale imitation data. Similarly, \cite{xue2025leverb} integrate VLA reasoning with primitive-based whole-body control, demonstrating strong semantic robustness but limited adaptability. 
\subsubsection*{Multi-modal Embodiment Datasets} 
Large-scale multi-embodiment datasets have played a central role in recent progress on VLA training. The Open X-Embodiment dataset~\cite{o2024open} aggregates over 1 million real-robot trajectories spanning 22 distinct robot embodiments, enabling vision–language reasoning and diverse manipulation tasks across heterogeneous hardware platforms. While such datasets have driven rapid advancements in manipulation-focused VLAs, recent success in this area has motivated early efforts toward collecting demonstration data for whole-body humanoid control. However, large-scale visual demonstrations for full-body behaviors remain scarce due to the difficulty of capturing rich whole-body motions on physical humanoid robots.
Several workarounds have been explored. LeVERB~\cite{xue2025leverb} provides synthetic whole-body demonstrations for the G1 humanoid robot, while Locomujoco~\cite{al2023locomujoco} offers a benchmark comprising motion-capture trajectories and expert rollouts but lacks aligned multi-modal data. HumanoidBench~\cite{sferrazza2024humanoidbench} introduces a simulation environment capable of generating sensory inputs for whole-body manipulation and locomotion, though it does not supply ready-to-use, aligned VLA training sequences. Similarly, SMPL-Olympics~\cite{luo2024smplolympics} focuses on physically simulated sports behaviors for humanoid characters rather than multi-modal robot demonstrations. Mimicking-Bench~\cite{liu2024mimicking} provides datasets for learning humanoid whole-body interactions by imitating human motion patterns; however, it lacks natural-language annotations and is not publicly released.
More recent initiatives such as LangWBC and ToggleMimic~\cite{shao2025langwbc, zheng2025togglemimic} distill privileged, motion-aware teachers into text-conditioned controllers, illustrating how natural-language goals can directly drive whole-body humanoid behaviors via end-to-end language–action models. Despite recent advances, multi-modal datasets for whole-body humanoid control remain scarce. Existing resources often lack locomotion rich demonstrations, comprehensive multi-modal annotations, and standardized benchmarks for diverse whole-body behaviors. This gap highlights the need for an accessible,  high-quality and well-structured dataset that captures a broad range of locomotion tasks including more challenging behaviors such as stair climbing and supports scalable evaluation of humanoid control methods.
\section{Methodology} 
\label{sec:methodology}
\begin{figure*}[t]
    \centering
    \includegraphics[width=\textwidth,height=8cm]{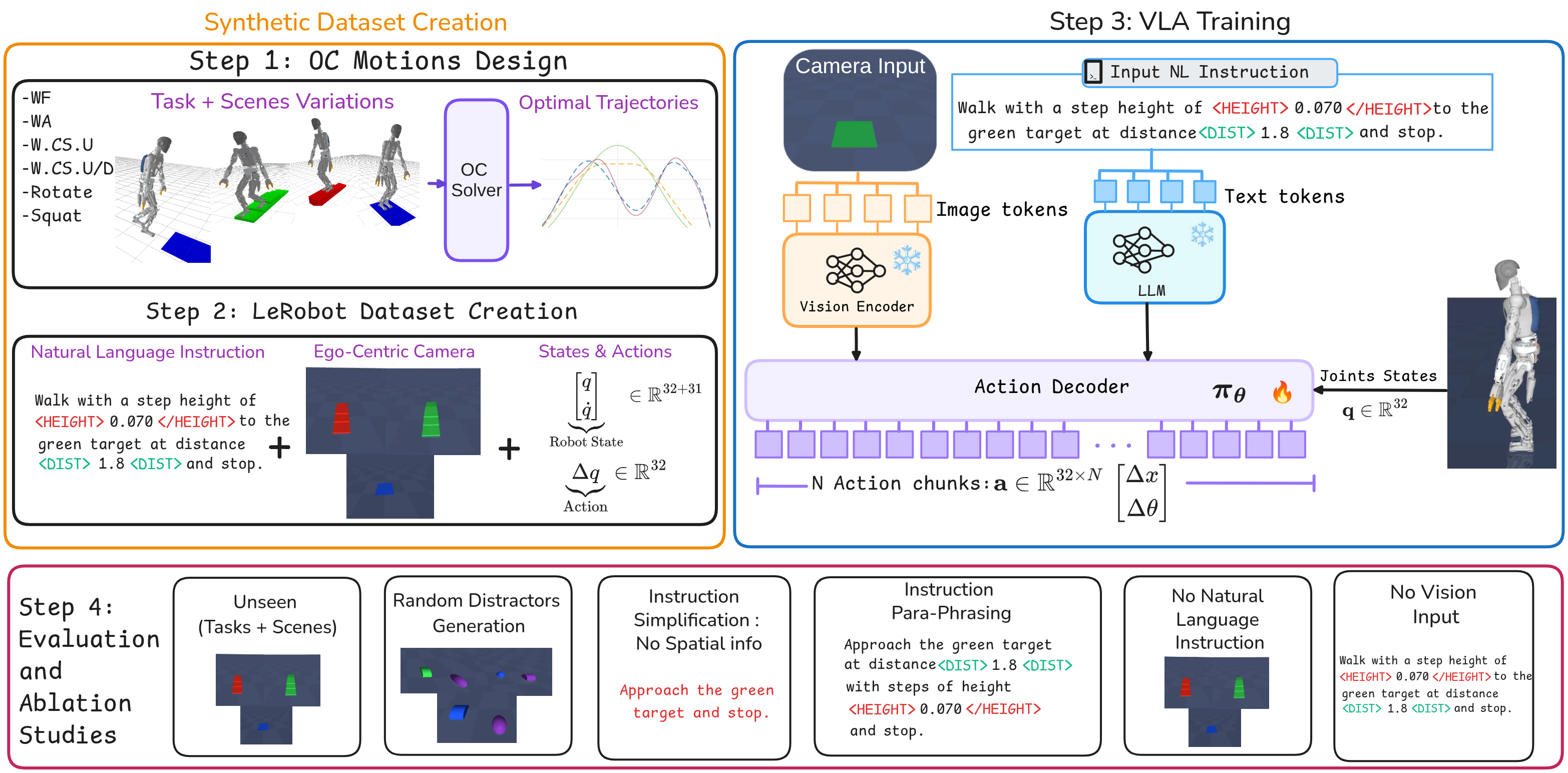}
    \caption{System pipeline of the proposed WOLF-VLA implementation.}
    \label{fig:pipeline}
    \vspace{-0.7cm}
\end{figure*}
\subsection{Synthetic Dataset Creation}
To address the lack of large-scale VLA datasets for whole-body humanoid control, we propose a scalable demonstration generation pipeline based on OC. Motion trajectories are generated by solving OCPs, producing dynamically consistent behaviors that minimize joint torques, rotations, and velocities while ensuring smooth contact transitions and strict satisfaction of joint constraints. This principled formulation yields physically consistent demonstrations and to the best of our knowledge enables the construction of the first large-scale dataset of whole-body humanoid locomotion behaviors generated through OC.

Randomized trajectory optimization across diverse environmental configurations is used to generate a wide range of locomotion behaviors, as detailed in \tref{tab:dataset_summary}. The resulting dataset contains 277 hours of humanoid motion spanning six task families with extensive parametric variations. These families correspond to distinct locomotion skills:
\begin{itemize}
    \item Target-directed locomotion: forward walking toward a target.
    \item Lateral locomotion: left and right side-walking toward a target.
    \item Stair interaction: forward walking followed by ascending a three-step staircase.
    \item Compound stair behaviors: forward walking with stair ascent and descent.
    \item $180^\circ$ turning motions, and Variable-height squatting.
\end{itemize}
\paragraph*{Environment Settings.}
As depicted in \fref{fig:open}, the environment is varied along three dimensions: object appearance, spatial configuration, and scene complexity:
\begin{itemize}
    \item Target types: box, cylinder, sphere, ground marker, single staircase (three steps), and double staircase (three steps up and down).
    \item Color variants: six colors applied to all target types.
    \item Spatial configuration: systematic variation along the $X$ and $Y$ axes, yields approximately $40 \times 40$ placements.
    \item Distractors: randomly placed non-target objects to increase visual diversity.
\end{itemize}

OC trajectories are executed in the MuJoCo physics simulator~\cite{todorov2012mujoco}, where time-synchronized visual observations are recorded. Videos are captured at 33.33\,Hz and RGB frames are extracted as visual inputs. In parallel, NL instructions describing each task are automatically generated. Each dataset sample therefore consists of RGB observations, the corresponding OC trajectory, and a task-level NL instruction. This multi-modal structure forms the WOLF-VLA dataset and enables VLA learning for whole-body humanoid locomotion rich. The overall pipeline is illustrated in \fref{fig:pipeline}.
\subsection{Action Skills Expressed as an OCP}
\paragraph*{Dynamics Under Contacts} To construct the motion dataset, we rely on the OC formalism of dynamically consistent motions in order to exploit the inherent optimality of motion and energy that can be embedded within the formulation. For this purpose, we model the robot dynamics as a multi-body system subject to $K$ contact constraints, following the Euler--Lagrange equations of motion (\ref{eq2}).
\begin{equation}
  \Mass(\q) \ddq + \Bias(\q,\dq) = \mathtt{S}^\intercal \bm \Torque + \sum_{k=1}^K \Jac_{k}(\q)^\intercal \bm \Force_k 
  \label{eq2}
\end{equation}
Here, $\q$ denotes the generalized joints configuration (internal joints + free-flyer),  $\dq$ are the generalized joints velocities and $\Mass(\q)$ represents the inertia matrix. The term $\Bias(\q,\dq)$ gathers the Coriolis, centrifugal, and gravity effects. The matrix $\mathtt{S}$ is the actuation selection matrix, while $\Torque$ denotes the vector of joint torques. Let $K$ be the number of contacts; for each contact $k$, $\Jac_{k}(\q)$ is the Jacobian of the $k^{\text{th}}$ contact and $\Force_{k}$ is the corresponding contact force. See~\cite{Fea14} for details.
During locomotion, the body maintains contact through one or both feet; accordingly, these contacts are modeled as second-order kinematic constraints on the contact placement in the equation of motion \eref{eq2}, since the dynamics is formulated in the acceleration space \cite{MelyaLift}.
\paragraph*{Action Optimality}
Consider humanoid locomotion relying on the multi-body dynamics under contacts in \eref{eq2}, formulated as a multi-phase OCP in which each phase corresponds to a specific contact configuration.
\footnote{$\q$, $\dq$, $\ddq$, $\Torque$ are functions of $t$. We drop the dependence here for clarity.}:
\begin{equation}\label{eq:ocpgen_wb}
\begin{aligned}
\operatorname*{minimize}_{\q,\dq,\Torque}
&\quad {\mathlarger{\sum}_{s=1}^S}\!\int_{t_s}^{t_s+\dts}
\!\costl_s(\q,\dq,\Torque)\,dt \quad \st \\
& \q\in\mathcal{Q},\;\dq\in\mathcal{V},\;\Torque\in\mathcal{T} \\
& \BIN\dq,\ddq\BOUT^\intercal
= f(\q,\dq,\Torque)
\end{aligned}
\end{equation}
where, $s=1 \cdots S$ is used to describe the contact phases which define the locomotion, $l_s$ is the cost function for the contact phase $s$, $f$ is the contact constrained dynamics \eref{eq2}, and $\mathcal{Q,V,T}$ are the admissible limits for $\q,\dq,\Torque$ defined by the motors.
The multi-phase OCP is solved via a multiple shooting formulation, the Differential Dynamic Programming (DDP)~\cite{Die17}, exploiting the sparsity of the Markovian dynamics in \eref{eq2}.
We use the open-source framework Crocoddyl~\cite{mastalli2020direct}, based on Pinocchio for efficient dynamics and derivatives~\cite{pinocchioweb}, and employ its Box-FDDP variant~\cite{mastalli2020direct} to enforce torque limits.
\paragraph*{Skills Design}
Solving the OCP for each motion yields the optimal trajectories of the state $x = (\q,\dq)$ and the corresponding optimal torque control $u = \Torque$.
To construct the dataset, we focus on variations of humanoid whole-body locomotion, including forward walking, lateral walking, and stair climbing.
The cost function structure is identical across all motion phases, differing only in the specified target references. The cost terms used are listed below:
\begin{equation}
  \costl_s = {\mathlarger{\sum}_{n=1}^N} \alpha_n \Phi_n(\q,\dq,\Torque),
\end{equation}
with cost terms $\Phi_n$ and empirically tuned weights $\alpha_n$, where $t_s$ denotes the final time per phase and $N$ the number of discretization steps per phase.
\begin{itemize}
  \item \textit{CoM Cost}: the CoM trajectory $c(t)$ tracks the target CoM position at the end of each phase, $c^{ref}(t_s)$. \\
    $\Phi_{1} = \parallel c(t)-c^{ref}(t_s)\parallel^{2}_2 $
  \item \textit{Feet Cost}: the stepping foot position trajectory $r(t)$ tracks the target step position per phase, $r^{ref}(t_s)$. \\
    $\Phi_{2} = \parallel r(t)-r^{ref}(t_s)\parallel^{2}_2$
  \item \textit{Torque minimization}: to guarantee dynamic feasibility.
    $\quad \Phi_{5} = \parallel \Torque (t) \parallel^{2}_2, $
  \item \textit{Posture regularization}: to handle the redundancy of the robot model, using the default robot posture $\q^{def}(t_s)$ as reference. 
    $\Phi_{6} = \parallel \q(t)-\q^{def}(t_s)\parallel^{2}_2$
\end{itemize}
Additionally to the automatique OCP resolution success check, each motion was manually validated for success, minimizing numerical errors and preserving dataset quality.
\subsection{Virtual Environment}
Our virtual environment is based on the gymnasium~\cite{towers2024gymnasium} simulation framework. All demonstrations and evaluations are conducted using the whole-body humanoid robot RH5 \cite{jin2025evolutionarycontinuousadaptiverlpowered,MelyaLift}. The RH5 model includes a free-flyer and 25 actuated joints, enabling rich whole-body locomotion behaviors. Our VLA benchmark adopts a multi-modal observation space that integrates both proprioceptive states, visual information and a NL instruction of the task. The proprioceptive state includes the full humanoid kinematic configuration, consisting of 32 joint rotations and 31 velocities, including the floating base. Formally, the proprioceptive observations at time $t$ is given by $\mathbf{o}^{\mathrm{prop}}_t = (\mathbf{q}_t, \dot{\mathbf{q}}_t)$, where $\mathbf{q}_t$ contains both actuated joint angles and the free-flyer pose parameters. Additionally, building on real-world vision-based WBC, we attached the humanoid head link with an ego-centric virtual camera  with a 120 degrees field of view in order to provide a time aligned 224 $\times$ 224 RGB image $\mathbf{o}^{\mathrm{img}}_t$ representing the visual observation of the environment. The action space is defined as a delta joint rotation command, $\mathbf{a}_t = \Delta \mathbf{q}_t$, representing incremental target updates to the actuated joint rotations at each control step. This formulation supports stable whole-body control while enabling direct learning of complex locomotive behaviors from  vision-language-proprioception inputs.
\subsection{Natural Language Instruction}
The generation of task instructions is fully automated and produced in parallel with the OCP resolution, leveraging the metadata recorded for each OCP formulation. Each NL instruction is augmented with structured tags that encode task-relevant information. 
Specifically, spatial position or distance to the target object is enclosed within \textcolor{BrickRed}{\texttt{\textbf{<DIST>}}} and \textcolor{BrickRed}{\texttt{\textbf{</DIST>}}} tokens, and for tasks in which the stepping height varies from the default one
the height information is wrapped within \textcolor{Emerald}{\texttt{\textbf{<HEIGHT>}}} and \textcolor{Emerald}{\texttt{\textbf{</HEIGHT>}}} tokens. During VLA model training, these tokens explicitly guide the LLMs component to better exploit the task spatial information within the NL instruction.
\subsection{VLA Policy Learning}
To learn a unified VLA policy from our constructed dataset, we train a transformer-based VLA model that jointly encodes visual observations, NL task instructions, and low-level joint trajectories. The training strategy follows a sequence-modeling formulation, where action prediction is cast as next-token auto-regressive modeling over continuous or discretized action representations. We train a VLA model that comprises three main components:  a visual encoder $\mathbf{E}_\mathrm{vis}$ that maps frame sequences into spatio-temporal tokens;  a language encoder $\mathbf{E}_\mathrm{lang}$ that embeds the NL task instruction; and  an action decoder $\mathbf{D}_\mathrm{act}$ that auto-regressively predicts control actions. Formally, Let $o$ denote the observation (vision, language, proprioception),
and let $a \in \mathbb{R}^{H \times d}$ be a ground-truth action chunk
of horizon $H$ and action dimension $d$. Following~\cite{black2024}, we sample Gaussian noise $\epsilon \sim \mathcal{N}(0, I)$ and a time
parameter $\tau \sim \mathcal{U}(0,1)$. A noisy interpolation between
noise and action is defined as:
\begin{equation}
a^{\tau} = \tau a + (1 - \tau)\epsilon .
\end{equation}
The model predicts a conditional vector field $v_{\theta}(a^{\tau} \mid o)$, which is trained to match the target flow direction $a - \epsilon$. The flow-matching objective is:

\begin{equation}
\mathcal{L}_{\mathrm{fm}}(\theta)
=
\mathbb{E}_{o,a,\epsilon,\tau}
\left[
\left\|
v_{\theta}(a^{\tau} \mid o)
-
(a - \epsilon)
\right\|^2
\right].
\end{equation}
This objective encourages the model to learn a vector field that transports samples from a Gaussian prior toward the data distribution of expert actions conditioned on observations. During inference, action chunks are generated using $K$ denoising steps. We first sample an initial latent action from a Gaussian prior $a^{0} \sim \mathcal{N}(0, I)$ and then use forward Euler integration with a step size $\Delta \tau = 1/K$ to iteratively generate the action chunk, updating as follows:
\begin{equation}
a^{\tau_{k+1}}
=
a^{\tau_k}
+
\Delta \tau \;
v_{\theta}(a^{\tau_k} \mid o)
\qquad
\end{equation}
The final denoised action chunk is
$\hat{a} = a^{\tau_K}.$

\begin{table}[t]
\centering
\small
\setlength{\tabcolsep}{10pt}
\renewcommand{\arraystretch}{1.2}
\resizebox{0.75\columnwidth}{!}{
\begin{tabular}{lcc}
\toprule
\textbf{Task} & \textbf{\# Episodes} & \textbf{Avg. episode length(s)} \\
\midrule
WF & 2874  & 13.5 \\
WA & 8234 & 43.2 \\
W.CS.U & 2358 & 21.6 \\
W.CS.U/D & 1810 & 33.6 \\
\midrule
\textbf{Total} & \textbf{15276} & \textbf{28} \\
\bottomrule
\end{tabular}}

\caption{WOLF-VLA dataset episode statistics.}
\label{tab:dataset_summary}
\vspace{-0.7cm}
\end{table}
\section{Experiment Setup}
\label{sec:implementation}
\subsection{Implementation Details}
Our main model is initialized from the pretrained GR00T-N1.5-3B~\cite{black2024} VLA policy with the LLM and vision backbone encoders kept frozen during training while the action diffusion model and the projector layers are fully optimized.
We train on 4 NVIDIA A100 GPUs for 200{,}000 gradient steps using an effective batch size of 128 (32$\times$4). Following a warmup of 500 steps, we apply a cosine decay learning-rate schedule that peaks at $1\!\times\!10^{-4}$ and decays to $1\!\times\!10^{-5}$. Optimization is performed with AdamW using $\beta_1 = 0.95$, $\beta_2 = 0.999$, $\epsilon = 10^{-8}$, and a weight decay of $1\!\times\!10^{-5}$. We apply gradient clipping with a maximum norm of 10. All training runs use bfloat16  precision. We formatted the proposed dataset in LeRobot \cite{cadene2026lerobot} dataset format which helps standardizing experiments for future imitation learning research work.

\subsection{Baselines and Ablation Settings}
We compare our approach against two representative policies  $\pi_{0.5}$\cite{intelligence2025pi05visionlanguageactionmodelopenworld} and
ACT\cite{zhao2023learningfinegrainedbimanualmanipulation} trained using the same observation and action representation space as our method to ensure a fair comparison.
We perform ablation studies to analyze the contribution of the different modalities used in our framework. The evaluated configurations are summarized in \tref{table:success_only_ablations}.

\noindent\textbf{GROOT-N1.}
Full model using ego-centric vision, proprioception, and natural language instructions with spatial cues.

\noindent\textbf{w.o spatial information.}
Spatial tags are removed from the language instructions to evaluate the importance of spatial grounding.

\noindent\textbf{w.o language.}
Natural language inputs are removed, forcing the policy to rely only on visual and proprioceptive observations.

\noindent\textbf{Instruction paraphrasing.}
paraphrased versions of the instructions generated using GPT-4o are used during evaluation to test robustness to linguistic variation.

\noindent\textbf{w.o vision.}
Visual observations are removed, leaving only proprioception and language inputs.


\subsection{Metrics and Evaluation Protocol}
All configurations are evaluated using the same protocol with 20 rollouts per task under both nominal and disturbed environments, and across increasing horizon lengths: short (S), medium (M), and long (L). 
Although individual target objects and colors appear separately in the training dataset, their combinations are excluded from the evaluation dataset.
We report the average success rate for each task and conduct ablation studies to analyze the effect of environment settings and task horizon lengths.
We evaluate the proposed dataset and learned policies using the success rate, and the ROM error ($\Delta$ROM), defined as follows.

\paragraph{Success Rate}
Performance is evaluated using a binary success rate. A rollout is successful if the robot completes the task objective within the allotted horizon. For $M$ rollouts. The reported success rate applies to all tasks except the stair-climbing tasks W.CS.U and W.CS.U/D.
\paragraph{Soft Success Rate (SSR)}
For a finer evaluation of VLA models ability to solve the compositional task of stair-climbing tasks (applied only for: W.CS.U + W.CS.U /D), we define a soft success rate that quantifies partial task completion proportionally to the number of reached stairs. Let a stair climb or descent task consist of $N$ discrete stairs representing an intermediate objective. SSR is defined as in \eref{eq:SSR}, the ratio 
between the number of reached intermediate objectives and the total number of objectives: 
\begin{equation}
\label{eq:SSR}
\mathrm{SSR} = \frac{1}{N} \sum_{i=1}^{N} s_i, 
\end{equation}
where $s_i = 1$ if the robot successfully reaches a stair $i$, and $s_i = 0$ otherwise.

\paragraph{ROM Error Rate ($\Delta$ROM)}
We evaluate how closely the learned policy reproduces the reference joint kinematics using the ROM of the robot joints.
The ROM of a joint rotation is defined as in \eref{eq:ROM}:
\begin{equation}
\label{eq:ROM}
\mathrm{ROM} = \left| \theta_{\max} - \theta_{\min} \right|.
\end{equation}

Let $t \in \{1,\dots,T\}$ denote the task index and 
$m \in \{1,\dots,M\}$ denote the joint metric. In this work, due to space limitations, we report only the most important joint rotations involved in locomotion tasks, namely the hip, knee, and ankle joints.
For each task and joint metric, the reference and actual ROM statistics are defined as:
\[
(\sigma^{(R)}_{t,m}, \mu^{(R)}_{t,m})
\quad \text{and} \quad
(\sigma^{(A)}_{t,m}, \mu^{(A)}_{t,m}),
\]
where $\mu$ denotes the mean ROM and $\sigma$ the corresponding standard deviation across steps.
The ROM error rate is then defined as the normalized difference between the mean actual and reference ROM values, scaled by the maximum reference ROM observed for that joint across all tasks:
\begin{equation}
\Delta \mathrm{ROM}_{t,m} =
\frac{\left| \mu^{(A)}_{t,m} - \mu^{(R)}_{t,m} \right|}
{\max_{t'} \mu^{(R)}_{t',m}}.
\end{equation}

This normalization ensures that the error reflects the relative deviation in joint utilization with respect to the largest reference motion range observed for each joint.


%
\begin{table*}[t]
\centering
\scriptsize
\setlength{\tabcolsep}{2pt}
\renewcommand{\arraystretch}{1.0}
\resizebox{0.99\textwidth}{!}{
\begin{tabular}{
l
scccscccscccscccscccsccc
s
}
\toprule
& \multicolumn{8}{c}{\textbf{S}}
& \multicolumn{8}{c}{\textbf{M}}
& \multicolumn{8}{c}{\textbf{L}}
& \textbf{Avg.} \\
\cmidrule(lr){2-9}\cmidrule(lr){10-17}\cmidrule(lr){18-25}
\textbf{Tasks}
& \multicolumn{4}{c}{\textbf{No Dist.}}
& \multicolumn{4}{c}{\textbf{Dist.}}
& \multicolumn{4}{c}{\textbf{No Dist.}}
& \multicolumn{4}{c}{\textbf{Dist.}}
& \multicolumn{4}{c}{\textbf{No Dist.}}
& \multicolumn{4}{c}{\textbf{Dist.}}
& \textbf{Succ.} \\
\cmidrule(lr){2-5}\cmidrule(lr){6-9}
\cmidrule(lr){10-13}\cmidrule(lr){14-17}
\cmidrule(lr){18-21}\cmidrule(lr){22-25}
& Succ. & {\tiny $\Delta$}HR & {\tiny $\Delta$}KR & {\tiny $\Delta$}AR
& Succ. & {\tiny $\Delta$}HR & {\tiny $\Delta$}KR & {\tiny $\Delta$}AR
& Succ. & {\tiny $\Delta$}HR & {\tiny $\Delta$}KR & {\tiny $\Delta$}AR
& Succ. & {\tiny $\Delta$}HR & {\tiny $\Delta$}KR & {\tiny $\Delta$}AR
& Succ. & {\tiny $\Delta$}HR & {\tiny $\Delta$}KR & {\tiny $\Delta$}AR
& Succ. & {\tiny $\Delta$}HR & {\tiny $\Delta$}KR & {\tiny $\Delta$}AR
& \\

\midrule
\multicolumn{9}{l}{\textbf{Groot-N1}} \\
\midrule
WF 
& 95.0 & 1.5 & 1.8 & 0.5 & 100.0 & 1.6 & 1.3 & 0.3 & 100.0 & 1.3 & 0.0 & 0.4 & 100.0 & 1.2 & 0.3 & 0.7 & 100.0 & 1.1 & 0.0 & 1.1 & 100.0 & 1.1 & 0.4 & 0.4 & 99 \\

WA
& 100.0 & 2.1 & 7.3 & 3.1 & 5.0 & 2.9 & 27.9 & 5.2
& 60.0 & 36.2 & 72.3 & 66.3& 0.0 & 100.1 & 344.9 & 74.8
& 0.0 & 23.4 & 312.3 & 117.8 & 0.0 & 47.0 & 33.1 & 83.8 & 27 \\

W.CS.U
& 55.0 & 1.3 & 1.1 & 1.1 & 46.7 & 1.5 & 1.6 & 1.5 & 48.3 & 1.3 & 1.7 & 1.3 & 63.3 & 1.3 & 2.0 & 1.0 & 48.3 & 1.3 & 0.8 & 1.1 & 50.0 & 1.3 & 1.3 & 1.6 & 51 (51) \\

W.CS.U/D
& 35.0 & 0.9 & 10.5 & 1.0 & 35.0 & 0.7 & 7.9 & 1.0 & 37.5 & 0.7 & 8.9 & 0.8 & 45.0 & 1.6 & 1.4 & 0.3 & 45.8 & 1.5 & 1.5 & 0.2 & 68.3 & 1.7 & 1.2 & 0.5 & 44 (10) \\

\midrule
\textbf{All}
& 71.3 & 1.5 & 5.2 & 1.4
& 46.7 & 1.7 & 9.7 & 2.0
& 61.5 & 10.0 & 21.8 & 16.4
& 52.1 & 54.1 & 339.4 & 38.0
& 48.5 & 51.5 & 489.8 & 160.3
& 54.6 & 39.0 & 367.5 & 78.9
 & 55.3 \\

\bottomrule
\end{tabular}}
\caption{\scriptsize
Performance results of different VLAs on the WOLF-VLA tasks under different ablation settings and horizon lengths: Short (S), Medium (M), and Long (L). For each task, we report the success rate (Succ. (\%)) over 20 episode runs and the average range of motion  error rate against ground truth trajectories for the most relevant joints orientation: hip ({\tiny $\Delta$}HR (\%)), knee ({\tiny $\Delta$}KR (\%)) and ankle ({\tiny $\Delta$}AR (\%)).Note that for climbing stair settings (W.CS.U, W.CS.U/D), \textbf{Succ.} refer to soft success rate and only the average standard success rate is reported between brackets.
The last column reports the average success rate across S, M, and L. The “All” row reports averages across all tasks. Tasks abbreviations stand for: WF = Walk Forward; WA = Walk Around; W.CS.U = Walk and Climb Stairs Up; W.CS.U/D = Walk and Climb Stairs Up and Down. 
}
\label{table:performance_results_restructured}
\vspace{-0.6cm}
\end{table*}

\begin{table}[t]
\centering
\scriptsize
\setlength{\tabcolsep}{3pt}
\renewcommand{\arraystretch}{1.0}
\resizebox{0.75\columnwidth}{!}{
\begin{tabular}{lccccccc}
\toprule
& \multicolumn{2}{c}{\textbf{S}} 
& \multicolumn{2}{c}{\textbf{M}} 
& \multicolumn{2}{c}{\textbf{L}} 
& \textbf{Avg.} \\
\cmidrule(lr){2-3}\cmidrule(lr){4-5}\cmidrule(lr){6-7}
\textbf{Tasks} 
& No Dist. & Dist. 
& No Dist. & Dist. 
& No Dist. & Dist. 
& Succ. \\

\midrule
\multicolumn{8}{l}{\textbf{ACT}} \\
All & 8.3 & 0.0 & 0.0 & 0.0 & 0.0 & 0.0 & 1.4 \\

\midrule
\multicolumn{8}{l}{\textbf{PI05}} \\
All & 0.0 & 0.0 & 0.0 & 0.0 & 0.0 & 0.0 & 0 \\

\bottomrule
\end{tabular}}
\caption{\small
Success rate (\%) of baseline VLAs on WOLF-VLA tasks under different ablations and horizon lengths: Short (S), Medium (M), and Long (L). Report averages over 20 episodes.}
\label{table:success_only}
\vspace{-0.1cm}
\end{table}

\begin{table}[t]
\centering
\scriptsize
\setlength{\tabcolsep}{3pt}
\renewcommand{\arraystretch}{1.0}
\resizebox{0.85\columnwidth}{!}{
\begin{tabular}{lccccccc}
\toprule
& \multicolumn{2}{c}{\textbf{S}} 
& \multicolumn{2}{c}{\textbf{M}} 
& \multicolumn{2}{c}{\textbf{L}} 
& \textbf{Avg.} \\
\cmidrule(lr){2-3}\cmidrule(lr){4-5}\cmidrule(lr){6-7}
\textbf{Tasks} 
& No Dist. & Dist. 
& No Dist. & Dist. 
& No Dist. & Dist. 
& Succ. \\
\midrule

\multicolumn{8}{l}{\textbf{Groot-N1}} \\
WF & 95.0 & 100.0 & 100.0 & 100.0 & 100.0 & 100 & 99 \\
WA & 100.0 & 5.0 & 60.0 & 0.0 & 0.0 & 0.0 & 27 \\
W.CS.U & 55.0.0 & 46.7 & 48.3 & 63.3 & 48.3 & 50.0 & 51 \\
W.CS.U/D & 35.0 & 35.0 & 37.5 & 45.0 & 45.8 & 68.3 & 44 \\

\midrule
\multicolumn{8}{l}{\textbf{Groot-N1 w.o spatial info}} \\
WF & 95.0 & 100.0 & 100.0 & 100.0 & 100.0 & 100 & 99 \\
WA & 85.0 & 5.0 & 30.0 & 0.0 & 0.0 & 0.0 & 20 \\
W.CS.U & 61.7 & 3.3 & 68.3 & 60.0 & 50.0 & 10.0 & 42 \\
W.CS.U/D & 31.7 & 4.2 & 51.7 & 50.0 & 49.2 & 4.2 & 31 \\

\midrule
\multicolumn{8}{l}{\textbf{Groot-N1 w.o language}} \\
WF & 95.0 & 100.0 & 100.0 & 100.0 & 100.0 & 100 & 99 \\
WA & 90.0 & 5.0 & 20.0 & 0.0 & 0.0 & 0.0 & 19 \\
W.CS.U & 53.3 & 0.0 & 51.7 & 48.3 & 55.0 & 21.7 & 38 \\
W.CS.U/D & 35.8 & 3.3 & 45.8 & 59.2 & 45.8 & 6.7 & 32 \\

\midrule
\multicolumn{8}{l}{\textbf{Groot-N1 + Instruction paraphrasing}} \\
WF & 95.0 & 100.0 & 100.0 & 100.0 & 100.0 & 100 & 99 \\
WA & 80.0 & 10.0 & 45.0 & 0.0 & 0.0 & 0.0 & 22 \\
W.CS.U & 68.3 & 35.0 & 51.7 & 50.0 & 55.0 & 48.3 & 51 \\
W.CS.U/D & 40.0 & 35.0 & 40.0 & 35.0 & 55.0 & 45.8 & 42 \\

\midrule
\multicolumn{8}{l}{\textbf{Groot-N1 w.o vision}} \\
WF & 20.0 & 15.0 & 0.0 & 0.0 & 0.0 & 0.0 & 5 \\
WA & 0.0 & 0.0 & 0.0 & 0.0 & 0.0 & 0.0 & 0 \\
W.CS.U & 23.3 & 21.7 & 8.3 & 8.3 & 0.0 & 5.0 & 11 \\
W.CS.U/D & 31.7 & 30.8 & 0.0 & 2.5 & 1.7 & 2.5 & 11 \\

\bottomrule
\end{tabular}}
\caption{\small
Success rate (\%) of VLAs on WOLF-VLA under different task ablations.}
\label{table:success_only_ablations}
\vspace{-0.7cm}
\end{table}

\section{Results and Discussion}
\label{sec:results}
Performance is measured under two environmental conditions: inference execution without visual distractors and with randomly injected distractors. This setup allows us to evaluate both task completion capability and robustness to perturbations.

\tref{table:performance_results_restructured} summarizes the results across all tasks. Tasks are grouped by horizon length reflecting progressively longer sequences of actions and increasing environmental complexity. In addition to task success rates, we report joint-level range-of-motion errors with respect to unseen OC reference trajectories. Specifically, we measure $\Delta$ROM errors for the hip, knee, and ankle joints, which represent the primary joints responsible for locomotion. These metrics provide insight into the fidelity of the generated motions relative to the OC demonstrations.

\paragraph{Motion fidelity and locomotion quality}

Across successful executions, the observed $\Delta$ROM errors remain consistently low, indicating that the learned policy closely reproduces the joint trajectories present in the OC demonstrations. This result suggests that the model does not simply learn task completion through arbitrary control strategies, but instead captures structured locomotion patterns encoded in the optimal trajectories. In practice, this leads to stable and physically plausible motion generation, as the policy implicitly inherits efficient locomotion strategies from the OC-generated dataset.

\paragraph{Long-horizon task performance}

A key challenge in humanoid locomotion is maintaining stability and consistency over long sequences of actions. As shown in \tref{table:performance_results_restructured}, the proposed model maintains stable performance across increasing horizon lengths. For example, the WF task achieves near-perfect success rates across all horizons, indicating that the policy can reliably sustain locomotion over extended trajectories. More complex tasks, such as  stair climbing, exhibit lower success rates but remain solvable even at longer horizons. This behavior suggests that the model effectively captures the temporal dependencies required for long-horizon control.

\paragraph{Robustness to visual distractors}

Experiments further highlight the robustness of the learned policy. For simple locomotion tasks such as WF, the policy remains largely unaffected by distractors, maintaining success rates close to 100\%. For more complex tasks such as WA or stair climbing, distractors introduce additional perceptual and coordination challenges, leading to moderate decreases in success rate. Nevertheless, the policy remains capable of recovering from perturbations in a substantial portion of trials, demonstrating that the learned control strategies generalize beyond the nominal trajectories observed during training.

\paragraph{Comparison with baseline methods}

We compare the proposed approach against two representative baseline policies: ACT and $\pi_{0.5}$. As reported in \tref{table:success_only}, both baselines fail to achieve meaningful success rates across the evaluated tasks. We argue that pretraining data lack strong multimodal conditioning
Notably, these results are obtained even though the GROOT N1.5 backbone used in this work was not pretrained on humanoid whole-body locomotion data. This observation suggests that the architecture can effectively adapt to locomotion control when trained on the proposed dataset.

\paragraph{Role of multimodal inputs.}

The ablation experiments reported in \tref{table:success_only_ablations} provide insights into the contribution of each modality. Removing spatial tags from the natural language instructions results in only a minor reduction in success rates, indicating that spatial cues in language serve primarily as contextual guidance rather than a strict requirement for task execution. Similarly, removing language entirely leads to a moderate decrease in performance for complex tasks, suggesting that language provides useful high-level task information but is not the sole source of task grounding. In contrast, removing visual observations leads to a dramatic drop in performance across all tasks. Without ego-centric visual input, the policy struggles even with simple locomotion tasks and fails entirely on tasks requiring environmental awareness. This result highlights the critical role of vision in enabling the policy to interpret scene structure and adapt locomotion behavior accordingly. Importantly, the results also demonstrate that ego-centric visual observations alone provide sufficient environmental context to support long-horizon locomotion, making them a practical sensing modality for real-world humanoid systems.

\paragraph{Dataset scalability and transfer potential.}

Beyond the empirical performance improvements, the proposed dataset generation pipeline provides several advantages for scalability and reuse. Because the dataset is generated using an optimal control framework, additional locomotion behaviors can be synthesized by modifying task objectives or environment configurations. Furthermore, visual diversity can be increased by regenerating environments with varied layouts or photorealistic rendering, which could further improve robustness to domain variations. Another key property of the dataset is its embodiment-agnostic design. The visual observations do not encode robot-specific information, and the natural language instructions describe task objectives independently of a particular robot platform. As a result, the dataset can be reused to train policies for different humanoid robots with similar joint limits by retargeting the OC-generated trajectories to match the range-of-motion constraints of the target platform.

\section{Conclusion and Outlooks}
\label{sec:conclusion}
This work presented a unified framework for learning vision-language-action models for whole-body humanoid locomotion. The approach combines three components: a large-scale dataset generated through optimal control, a VLA model trained on this dataset, and a benchmarking suite with systematic ablations. Together, these elements provide a reproducible pipeline for studying VLA-based control for humanoid robots.
The dataset captures diverse locomotion behaviors across multiple task categories and horizon lengths, with trajectories generated through an OC formulation that enforces whole-body dynamics and contact feasibility. Experimental results show that models trained on this dataset achieve high success rates while producing motions that closely follow the OC reference trajectories. The learned policy maintains stable performance across both short- and long-horizon tasks and remains robust to environmental distractors.
Benchmarking results show that the large pretrained models consistently outperforms baseline approaches, suggesting that combining dynamically consistent OC demonstrations with VLA training provides a strong foundation for learning complex whole-body behaviors. Additionally, the experiments highlight the effectiveness of ego-centric visual observations for locomotion tasks.
Overall, this work bridges OC-based motion generation and large-scale VLA policy learning, providing a scalable benchmark for humanoid locomotion. Future work will focus on expanding the dataset, improving long-horizon multi-task capabilities, and enabling safe real-world deployment on humanoid robots.
\bibliographystyle{IEEEtran}
\begin{samepage}
\bibliography{references}
\end{samepage}
\end{document}